%% file: main_arxiv.tex
\documentclass[9pt,twocolumn,a4paper]{article}

\usepackage{geometry}
\usepackage{palatino}
\usepackage{amsmath}
\usepackage{amssymb}
\usepackage{graphicx}
\usepackage{foreign}
\usepackage{booktabs}

\usepackage[font=footnotesize]{caption}

\input{preamble}

\definecolor{cvprblue}{rgb}{0.21,0.49,0.74}
\usepackage[pagebackref,breaklinks,colorlinks,allcolors=blue]{hyperref}

\newcommand{\matthijs}[1]{{\color{magenta}[\textbf{Matthijs}: #1]}}

\usepackage{hyperref}

\usepackage[capitalize]{cleveref}
\crefname{section}{Sec.}{Secs.}
\Crefname{section}{Section}{Sections}
\Crefname{table}{Table}{Tables}
\crefname{table}{Tab.}{Tabs.}

\newcommand{\citet}[1]{\cite{#1}}
\newcommand{\citep}[1]{\cite{#1}}

\title{Lossless Compression of Vector IDs \\
for Approximate Nearest Neighbor Search}

\date{}

\author{
Daniel Severo \\
{\tt\small dsevero@meta.com}
\and
\hspace{0.2cm}
\and
Giuseppe Ottaviano \\
{\tt\small ott@meta.com}
\and
\hspace{0.2cm}
\and
Matthew Muckley \\
{\tt\small mmuckley@meta.com}
\and
Karen Ullrich \\
{\tt\small karenu@meta.com}
\and
Matthijs Douze \\
{\tt\small matthijs@meta.com}
}

\begin{document}
\maketitle
\input{sec/0_abstract}
\input{sec/introduction}
\input{sec/related-work}
\input{sec/background}
\input{sec/method}
\input{sec/experiments}

\input{sec/conclusion}

\clearpage
{
    \small
    \bibliographystyle{plain}
    \bibliography{main,vector_search}
}

\clearpage
\setcounter{page}{1}
\appendix
\input{sec/supplementary}

\end{document}

%% file: preamble.tex
\usepackage{amssymb}
\usepackage{bbm}
\usepackage{amsthm}
\usepackage{amsmath}
\usepackage{multirow}
\usepackage[table,dvipsnames]{xcolor}
\input{commands}

%% file: commands.tex
\newcommand{\defeq}{\triangleq}

\newcommand{\X}{\mathcal{X}}
\newcommand{\Naturals}{\mathbb{N}}

\newcommand{\Codes}{\mathbf{X}}
\newcommand{\code}{\mathbf{x}}

\newcommand{\friendlist}{\mathbf{e}}

\newcommand{\bitstream}{\mathbf{s}}
\newcommand{\tablehead}[1]{\underline{\textbf{#1}}}
\newcommand{\ivf}{\boldsymbol{\ell}}

\theoremstyle{remark}

\usepackage{mathtools}
\DeclareMathOperator{\expect}{\mathbb{E}}
\newcommand\given{\,\vert\,}

\DeclarePairedDelimiterX{\divergence}[2]{(}{)}{#1\,\delimsize\|\,#2}
\newcommand{\kl}{D_\mathrm{KL}\divergence}

\newcommand{\encode}{\mathrm{encode}}
\newcommand{\decode}{\mathrm{decode}}
\DeclarePairedDelimiter\ceil{\lceil}{\rceil}
\newcolumntype{g}{>{\columncolor[gray]{0.9}}c}

%% file: sec/0_abstract.tex
\begin{abstract}
Approximate nearest neighbor search for vectors relies on indexes that are most often accessed from RAM. 
Therefore, storage is the factor limiting the size of the database that can be served from a machine. 
Lossy vector compression, i.e., embedding quantization, has been applied extensively to reduce the size of indexes.
However, for inverted file and graph-based indices, auxiliary data such as vector ids and links (edges) can represent most of the storage cost. 
We introduce and evaluate lossless compression schemes for these cases. 
These approaches are based on asymmetric numeral systems or wavelet trees that exploit the fact that the ordering of ids is irrelevant within the data structures.
In some settings, we are able to compress the vector ids by a factor 7, with no impact on accuracy or search runtime.
On billion-scale datasets, this results in a reduction of 30\% of the index size. 
Furthermore, we show that for some datasets, these methods can also compress the quantized vector codes losslessly, by exploiting sub-optimalities in the original quantization algorithm.
The source code for our approach available at~\url{https://github.com/facebookresearch/vector_db_id_compression}
\end{abstract}

%% file: sec/introduction.tex
\section{Introduction}\label{sec:intro}
\input{figs/fig-intro}
Vector search is at the foundation of most methods for the retrieval of images, videos or other media~\cite{sivic2003video,philbin2007object,weyand2020google,pizzi2022self}. 
Given embeddings vectors for a collection of media items, and a query embedding, the retrieval consists in finding the nearest vector from the collection, in terms of Euclidean, cosine, or other vector distances.
When the size of the collection grows, exact search becomes too slow, so practitioners resort to approximate nearest neighbor search (ANNS) that trades a large reduction in search time or memory usage for some loss of accuracy.  

Mathematically, a vector database can be considered as an $N\times D$ matrix whose rows are the $N$ embedding vectors in dimension $D$.
Then, given a query vector, the goal of nearest-neighbor search is to find the index of the vector from the database closest to the query.

We consider the task of compression for vector databases.
Compression of such databases gives benefits not only to the standard aspects of storage and transmission costs, but also the overall bandwidth of the search pipeline. 
For ANNS, there are two types of data to compress: the vectors themselves and sets of vector ids that are used in fast search structures. 
Vector compression has been studied extensively as a quantization task and is frequently leveraged to reduce the number of comparisons necessary for ANNS, but relatively little attention has been devoted to entropy encoding applied on the generated codes. 
Furthermore, fast vector search data structures also rely on sets of vector ids that may also benefit from lossless compression.

One indexing structure stores vectors and ids in clusters to enable a two-stage ANNS (\Cref{fig:intro}, top).
First, a number of clusters is chosen based on the distance of the query to the cluster centroids.
Exhaustive search is then performed in the union of clusters, implying the order between vectors and ids within a cluster is irrelevant.
A similar invariance of the edges in graph-based indices also exists (\Cref{fig:intro}, bottom).
We exploit this invariance to reordering to improve compression by storing vectors and ids in a set, or multiset, data-structure.
Shannon's information theory \cite{shannon1948, cover1991} establishes that representing a set of $n$ elements requires $\log n!$ \emph{less} bits (order information) than a sequence of the same elements.
This quantity can be significant for ANNS applications, where each cluster contains thousands of elements.

\paragraph{Orderless compression.}
\citet{severo2022compressing, severo2023random} presented compression methods for set, multisets, and graphs that can achieve the information-theoretic savings of the ordering.
As such, our work is dedicated to extending these methods for the case of ANNS.
We distinguish between two settings: \emph{offline} and \emph{online}.
In the offline setting, the index as a whole is compressed and decompressed. 
This is the case when it is stored or transmitted as a binary blob; it is not used before being decompressed as a whole. 
Therefore it is evaluated purely in terms of compression ratio.
In the online setting, the compressed index is in RAM and part of it is decompressed at search time when needed.
This is a more constrained setting, where the compression rate has to be traded off against the search speed.
A special case of the online setting is that of \emph{random access}: meaning it is possible to decompress an individual element of the sequence.

The contributions of this work are: 
\begin{itemize}
\item 
    We provide the first evaluation in terms of speed vs. compression rate of lossless compression applied to sets of vector ids for \emph{online} search, 
\item 
    We develop codecs based on asymmetric numeral systems (ANS) to compress the ids used in partitioning and graph vector search indexes;    
\item 
    We provide a graph compression algorithm applied to graph search in the \emph{offline} setting.    
\end{itemize}

After a review of related work in Section~\ref{sec:related}, Section~\ref{sec:background} provides the necessary background on the compression of sets of integers and graphs. 
In Section~\ref{sec:method} we show how to apply this to the compression of ANNS indexes. 
Sections~\ref{sec:experiment} evaluates these methods in the offline and online settings.

%% file: figs/fig-intro.tex
\begin{figure}
    \centering
    \includegraphics[width=\linewidth]{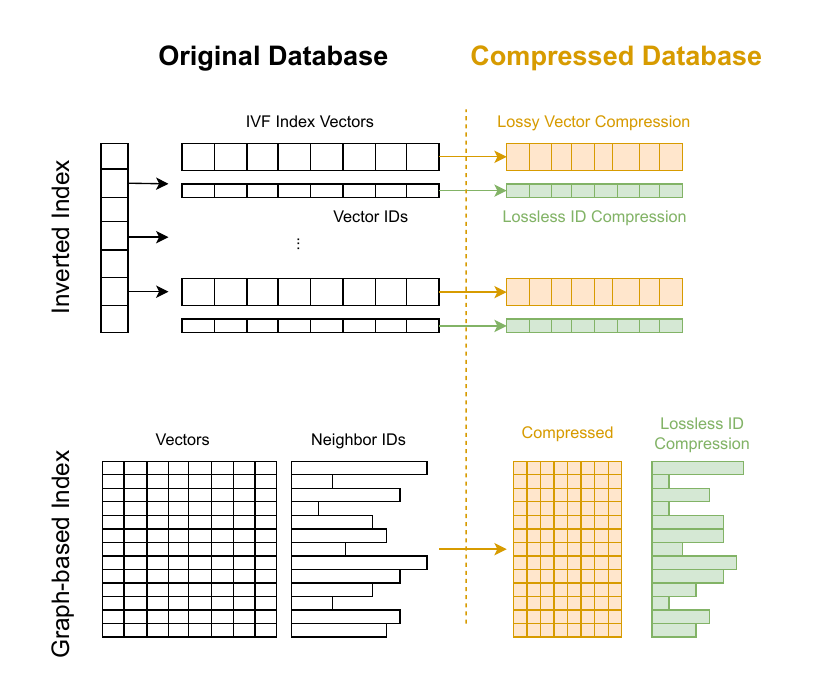}
    \caption{
        How we apply lossless compression in state-of-the-art indexes for vector search.
        (Orange) Embeddings/vectors are compressed using lossy techniques such as Product Quantization \cite{jegou2010product} and QINCo \cite{niu2023residual, huijben2024QINco}. (Green) We focus on lossless compression of identifiers (IVF, top) and links (edges in graph indices, bottom).
    }
    \label{fig:intro}
\end{figure}

%% file: sec/related-work.tex
\section{Related Work}
\label{sec:related}

\paragraph{Vector compression.}

The compression of images~\cite{wallace1992jpeg,rabbani2002overview} or videos~\cite{kwon2006overview} often proceeds by 
(1) lossless mapping of the continuous signal into a more favorable representation space,
(2) lossy compression of the data into a discrete sequence, and 
(3) lossless encoding of the discrete sequence.

A similar process takes place to compress vectors. 
Most works have studied the second step, which consists in a quantization of a vector to an integer of fixed size or a vector of bits: the code. 
Early quantization methods are simple: each vector component is quantized to one integer~\cite{gray1998quantization} in the extreme case a single bit~\cite{weiss2008spectral,jegou2008hamming}. 
More accurate vector quantizers~\cite{gray1998quantization} encode multiple dimensions simultaneously, using a vector codebook.
Practical vector quantizers exploit multiple vector codebooks to avoid codebooks sizes that grow exponentially with the code length. 
These quantizers include the Product Quantizer~\cite{jegou2010product}, additive quantizers~\cite{babenko2014additive,martinez2018lsq++} and quantizers based on neural encoder-decoder architectures~\cite{morozov2019unsupervised,huijben2024QINco}.

The first step, performing continuous-domain transformations of vectors has also been adjusted to the subsequent quantizer: ITQ for binary quantization~\cite{gong2012iterative}, optimized PQ for PQ~\cite{ge2013optimized} or a catalyzer for lattice quantizers~\cite{sablayrolles2018spreading}.

Step 3, which performs entropy coding to the encoded vectors, is not popular for vector encoding, due to the assumption that a good vector quantizer produces codes with maximum entropy (a uniform distribution) \cite{cover1991}.
The seminal work of~\citet{han2015deep} on neural net compression does apply Huffman encoding on quantized vectors, but it exploits a weak scalar quantizer that produces redundant codes. 
In this work, we apply state-of-the-art entropy coders to codes. 

\paragraph{Vector search}

To speed up searching, the data needs to be organized in a data structure that allows fast access to the most promising subset of vectors while ignoring the others: this is called search pruning.
The vector distances with the query are computed only for the pruned subset. 
Early methods were based on tree structures~\cite{muja2014scalable} and hashing~\cite{datar2004locality}. 
Current methods are based either on partitioning or graphs. 
Partitioning (a.k.a. inverted-file based, or IVF) consists in clustering the vectors of the collection. 
At search time only a subset of the clusters are visited~\cite{sivic2003video,jegou2010product,guo2020accelerating,sun2023automating}. 

Graph based approaches~\cite{dong2011efficient,malkov2018efficient,fu2017fast,subramanya2019diskann,ootomo2023cagra,chen2024roargraph} build a graph where nodes are the collection's vectors and edges represent navigation paths that are followed at search time to hop between nearest neighbor candidates.
The most classical graph indexing algorithm is HNSW~\cite{malkov2018efficient}. 
In this paper we focus on the NSG~\cite{fu2017fast} index that performs well relative to HNSW~\cite{douze2024faiss} and that has a simpler, non-hierarchical, graph structure. 
Interestingly, graph search and IVF are often combined for large datasets by using the graph-based index to partition the database~\cite{Baranchuk_2018_ECCV,douze2024faiss}.

In both cases, a significant amount of memory is devoted to storing sets of vector ids (integers between 0 and $N-1$), see Figure~\ref{fig:intro}.
For partitioning methods, the vectors are reordered so their ids need to be stored explicitly, along with the vectors. 
For graph based methods, each outgoing edge is defined by the id of the target node. 
Depending on the type of index, these ids can represent a significant fraction of the storage~\cite{douze2018link,subramanya2019diskann}.

In many ANNS implementations, like Faiss~\cite{douze2024faiss,guo2020accelerating}, the ids are represented with 32 or 64-bit machine words. 
A basic improvement is to store them as $\lceil \log(N) \rceil$ bits~\cite{Vardanian_USearch_2022}.
This work aims at compressing the beyond that size, by exploiting their distribution and the degrees of freedom offered by the order of the ids.

A related study is the Link\&Code method~\cite{douze2018link} that shows that for a given memory budget, the compression of vectors and the number of outgoing edges in the HNSW search index should be balanced. 
However no attempt was made to compress the edge ids themselves.

%% file: sec/background.tex
\section{Background}
\label{sec:background}

This section, outlines the key techniques we use for lossless compression of relevant data types, including sets, multisets, and graphs.
Baseline methods such as Elias-Fano \cite{elias1974efficient} and Zuckerli \cite{versari2020zuckerli} are discussed in \Cref{sec:appendix-baselines}.

\paragraph{Notation}
The following is a shorthand for contiguous intervals of integer: $[a) \defeq \{0, \dots, a-1\}$.
The set of all binary strings, of all lengths, is denoted by $\{0, 1\}^\star$.
For a sequence $\ivf \defeq [\ell_1, \dots, \ell_N]$, we say $i \in \ivf$ if some element of $\ivf$ is equal to $i$.
All logarithms are base $2$.

\subsection{Lossless Compression}
Given an infinite i.i.d.\ sequence of discrete random variables $x_i \sim p_d$, with alphabet $\X \subset \Naturals$, the objective of lossless compression is to find a bijection (i.e., code) between the alphabet of the infinite sequence, $\X^\infty$, and $\{0, 1\}^\star$, such that the expected number of bits is minimized.
The expected length of any code, normalized by the number of symbols, is lower bounded by the entropy~\cite{cover1999elements}: $H(p_d)= \expect_{x \sim p_d}\left[-\log p_d(x)\right]$.

In practice, the true distribution $p_d$ is  unknown, so a model probability mass function (PMF) $p(x)$ is used instead. 
The minimal achievable expected length in this scenario~\cite{mackay2003information,cover1999elements} is the cross-entropy of \(p\) relative to \(p_d\), given by $\expect_{x \sim p_d}\left[-\log p(x)\right] \geq H(p_d)$.
Equality is achieved only if $p(x) = p_d(x)$ for all $x$.

\paragraph{Asymmetric Numeral Systems (ANS)}
ANS are model-based coders that achieve compression rates close to the cross-entropy, when paired with a probability model $p$ \citep{duda2009asymmetric}. 
ANS creates a code via a bijection between sequences and natural numbers $s \in \Naturals$ or equivalently their binary representation $\{0, 1\}^\star$.

Encoding a symbol $x$ into the ANS state $s$ yields a new state $s^\prime \geq s$.
This requires that the probability model be quantized to some precision $r \in \Naturals$, such that $\sum_{x \in \X} p_x = r$, where $p_x \in \Naturals$ is the quantized probability value such that the ratio $p_x/r$ approximates the probability $p(x)$.

Encoding is performed directly on the integer state with the following function,
\begin{align}
  \encode_p&(s, x) \defeq r \cdot \lfloor s / p_x \rfloor + c_x + s \bmod p_x,
\end{align}
where $c_x = \sum_{y < x} p_y$ is the quantized CDF value.
To encode a sequence of symbols $(x_1, \dots, x_n)$, ANS successively applies $s_i \defeq \encode_p(s_{i-1}, x_i)$, starting from some $s_0$. 
The binary representation is the final state $s_n$.
Decoding proceeds in reverse fashion, starting from the last encoded symbol. 
For this reason, ANS is often referred to as a stack-like entropy coder.
If $s^\prime \defeq \encode_p(s, x)$, then the previous state $s$, as well as the encoded symbol $x$, can be recovered with
\begin{align}
    x &= s^\prime \bmod r \\
    \decode_p(s, x) &\defeq p_x \cdot \lfloor s^\prime / r \rfloor - c_x + x = s
\end{align}

With slight abuse of notation, we note the size of the binary representation of $s$ as $\log s$. %
This quantity changes according to the probability of the symbols under the model:
\begin{equation}\label{eq:ans-approx}
  \log s' = \log s - \log p(x) + \epsilon.
\end{equation}
ANS is near-optimal in that the redundancy per operation is typically bounded by \(\epsilon \leq 2.2 \times 10^{-5}\) bits \cite{townsend2020, duda2009asymmetric}.
\Cref{eq:ans-approx} implies the ANS state will grow by approximately the cross-entropy on average, which is the best achievable rate by a fixed probability model $p$.

The following are useful facts about coding with ANS, which we make use of in upcoming sections.
\begin{enumerate}
    \item Implementing ANS encoding/decoding requires routines to compute the cumulative distribution function (CDF) of $p$ efficiently, as well as its inverse.
\item 
    Given a sufficiently large ANS state \(s\), symbols can be decoded using any distribution \(p'\), producing a random symbol \(x\) and reducing \(s\) by approximately \(-\log p'(x)\) bits.  
    This process is invertible by re-encoding the sampled symbols, allowing the original state to be recovered.
    This means \(s\) can serve as an invertible sampler with a reservoir of randomness. 

\end{enumerate}

\subsection{Bits-back for Combinatorial Objects}\label{sec:background-bitsback}

\paragraph{Bits-back Coding}
Bits-back~\citep{townsend2019practical,hinton1993keeping} is a general strategy for lossless compression of random variables \(x\) with a discrete latent variable model (LVM) that factorizes as $p(x,z)=p(z)p(x \given z)$.
For this type of model, using an ANS codec directly is not feasible if we can not marginalize to obtain $p(x)=\sum_z p(z)p(x \given z)$.
Bits-back enables entropy coding with an LVM if we have access to an approximation $q(z \given x)$ of the model posterior $p(z \given x) \defeq p(x \given z)p(z)/p(x)$.
Assuming  $p(z),p(x \given z)$ and $q(z \given x)$ are tractable in the sense of 1. (i.e., tractable CDF), we can use observation 2. to construct a codec:
\begin{enumerate} %
\item 
    We use $\decode_{q(z|x)}$ to sample a latent variable $z$, removing $-\log q(z|x)$ bits from $s$. 
\item  
    We then use $\encode_{p(z)}$ and $\encode_{p(x|z)}$, to encode the sampled latent $z$, as well as the observation $x$, adding $-\log p(x,z)$ bits to $s$.
\end{enumerate}
The first step (i.e., sampling $z$) decreases the ANS state, which can be interpreted as providing a saving in rate equal to $\expect_{(x, z) \sim p_d(x)q(z \given x)}\left[-\log q(z \given x)\right]$.
This is sometimes referred to as the ``free energy" of the model \cite{hinton1993keeping}.
Due to this savings, the number of bits a symbol \(x\) adds to the bit stream \(s\) is equivalent to the Negative Evidence Lower Bound (NELBO) under the LVM:
\begin{align}\label{eq:NELBO}
  &\expect_{z\sim q(z\given x)}\left[-\log p(x, z) + \log q(z \given x)\right]\\
  &= - \log p(x) + \kl{q(z \given x)}{p(z \given x)}.\notag
\end{align}
In expectation, the KL term implies \Cref{eq:NELBO} is an upper bound to the cross-entropy.
If the approximate posterior is exact (i.e., $q(z \given x) = p(z \given x)$, for all $x$ and $z$) then coding with bits-back will achieve the same rate as encoding with ANS under the marginalized model $p(x)$.

\paragraph{Initial bits issue.}
Bits-back encodes a sequence of symbols $x_i$ by sampling a sequence of latents $z_i \defeq \decode_{q(z \given x)}(s_{i-1}, x_i)$, and then encoding $(x_i, z_i)$ with the LVM using the joint $p(x_i, z_i)$.
The initial state $s_0$ must be filled with a few random bits because the encoding process starts by \emph{decoding} $z_1$ from $s_0$,
i.e., some value for $s_0$ must be chosen to initiate the chain.
The initial bits are a one-time overhead in encoding a sequence, and will be amortized for large sequence lengths.

\paragraph{Random Order Coding (ROC)}
A sequence can be viewed as a set of elements together with a permutation defining the ordering (relative to some fixed, canonical ordering).
Conversely, an observed set $x$ can be represented by a sequence by interpreting the ordering as a latent variable $z$.
Bits-back can be utilized to compress sets and multisets as well as other combinatorial objects, by viewing these data types as sequences with latent permutations \cite{severo2022compressing}.

\citet{severo2022compressing} derived an exact expression for the posterior of an LVM with latent permutations. 
The
\emph{Random Order Coding} (ROC) codec is a practical implementation of bits-back coding for sets and multisets. 
The implementation takes care of setting the initial bits, and interleaves encoding $z$ and decoding $x$ to bound the memory requirements. 

\paragraph{Random Edge Coding (REC)}
\citet{severo2023random} extends the above ideas for compressing  graph data types.
A directed graph $x$ with $n/2$ edges can be represented by a sequence of nodes $v(x) \defeq (v_1(x), \dots, v_n(x))$, where an edge is encoded in the sequence by every 2 consecutive elements; $e_i = (v_{2i-1}, v_{2i})$.
The canonical sequence in this case must be constructed at the edge-level by sorting pairs $e_i$.
The directed graph can be represented by any of of the $\frac{n}{2}!$ sequences $(e_{z_1}, \dots, e_{z_n})$.

Similar to ROC, Random Edge Coding (REC) \cite{severo2023random} provides an efficient implementation to compress directed, undirected, as well as hypergraphs, while mitigating the initial bits problem.

\subsection{Wavelet tree codecs}
\label{sec:wavelettree}

We use a wavelet tree to index a sequence of symbols $S\in [K)^N$~\cite{grossi2012wavelet,bowe2010multiary}. 
The binary tree is built on a recursive subdivision of the vocabulary $[K)$. 
Each internal node is labeled by a subset of $L \subset [K)$, that is split into two subsets to form the two child nodes. 
The root of the tree has $L=[K)$ and the leaves have $|L| = 1$.

The string $S$ is split similarly over the tree: a node with subset $L$ keeps only the symbols of $S$ that are in $L$: $S_{|L}$. 
The node does not store $S_{|L}$ explicitly.
Instead it stores a binary string of size $|S_{|L}|$ that indicates whether each symbol of $S_{|L}$ is assigned to the the left or right child of the node. 
Note that if $K$ is a power of two, the union of these binary strings has the same size as the original representation: $N\log(K)$ bits.

The \verb|select| operation recovers the index in $S$ of the $\mathcal{O}$\textsuperscript{th} occurrence of symbol $k$. 
For this, it first descends to leaf $k$.
Then, when backtracking to the root, it needs to perform a series of $\log K$ \verb|rank| operations on binary strings, which finds the number of 0s before an occurrence of 1 in a binary string. 
This can be done in constant time with a specialized structure (the RRR). 

%% file: sec/method.tex
\section{Compressing Database Indices}
\label{sec:method}

This section discusses the application of index compression under $3$ scenarios, differing by setting: offline, online or full random access. %
The potential reduction in the number of bits required to represent the database is greater as the requirement is loosened.

\paragraph{Problem setup}
Let $\Codes \in [M)^{N \times d}$ be the matrix of code-words, resulting from some vector quantization algorithm, where $N$ is the database size. 
The code size is $d\log M$ (e.g., PQ16 \citep{jegou2010product} would have $d=16, M=256$). 

For IVF indices, the rows $\code_i$ of $\Codes$ are partitioned into $K$ non-overlapping sets to perform approximate nearest neighbor search.
The partition information is stored in the form of $K$ inverted lists, $\ivf_k \in [N)^\star$, such that: if $i \in \ivf_k$, then the $i$-th row of $\Codes$ is in partition $k$.
For graph indices, we take $\friendlist_i \in [N)^{m_i}$ to be a sequence of indices such that if $j \in \friendlist_i$, then the directed edge $\code_i \rightarrow \code_j$ is present in the graph; referred to as the \emph{friend list} of vector $i$. %
The vertex set is equal to $\Codes$.
\input{figs/fig-pq-walltime}
\paragraph{Exploiting invariances for compression.}
For the purpose of similarity search, both database types exhibit invariances with respect to the ordering of codes, order of elements in friend lists, as well as remapping of cluster indices.
More precisely, for any permutations $\sigma$ and $c$, over $[N]$ and $[K]$ respectively, we can permute the rows of $\Codes$ (i.e., $\code_i \rightarrow \code_{\sigma(i)}$), as well as partition \emph{labels} (i.e., $\ivf_k \rightarrow \ivf_{c(k)}$) 
without affecting search results or speed.
Similarly, for graph based indices the invariance is with respect to the order of elements in the friend list.
For the $i$-th friend list, we can apply a permutation $[m_i]$ to its elements without affecting search.

In this work, we exploit these invariances to improve lossless database compression.
The set of codes $\Codes$ is the result of applying vector quantization algorithms to the embedding vectors, and therefore we expect them to be mostly incompressible \cite{cover1999elements} (this is further investigated in \Cref{sec:compressing-codes}.)
For these reasons, we focus on methods for compressing the inverted lists $\ivf_k$ and friend lists $\friendlist_i$.
To achieve these savings, we employ algorithms that compress strictly monotone sequences of unique elements or, equivalently, sets. 
This includes Elias-Fano (\Cref{sec:elias-fano}) and ROC (\Cref{sec:background-bitsback}). 
These data types are equivalent in the following sense: for a fixed sequence length $N$, and alphabet from which elements are drawn, the number of distinct strictly monotone sequences, and the number of distinct sets that can be constructed, are equal \cite{severo2022compressing}.
In other words, imposing the requirement that a sequence be sorted is equivalent to interpreting that sequence as representing a set composed of its elements.

\paragraph{Potential savings in bits.}
The theoretical savings from these algorithms, measured in bits, can be computed by counting permutations as in \cite{severo2024randompermutationcodeslossless}: for a sequence of size $n$ this is $\log n! \approx n\log n$.
For IVF, if $N_k$ is the number of elements in the $k$-th cluster, then each cluster contributes a potential saving of $\log(N_k!)$ bits.
For graph-based methods, each friend list contributes $\log(m_i!)$ bits.
The gain due to the invariance of cluster relabeling is $\log(K!)$, and is only significant when $K \approx N$, as shown in \cite{severorandom}, which is impractical.
We follow the ANNS setting of \cite{johnson2019billion} which sets $K = a\cdot\sqrt{N}$, for some $a \ge 1$, implying the savings from cluster relabeling are small and can be ignored.
Note that for ANS-based methods, the saved bit amounts are close to the theoretical ones. 

\subsection{Full random access}

IVF search can benefit from full random access.
At IVF search time, a structure that keeps the top-k results (typically a priority queue or a sorting network~\cite{douze2014yael,johnson2019billion}) is maintained. 
Each distance computation proceeds as follows: 
(1) the distance between the query vector and the code $\code_i$ is computed; 
(2) this distance is compared with the current worst result in the top-k structure;
(3) if the distance if better than the worst, it is evicted to make space for the new vector id.

The ids for most vectors that are compared to the query are not needed at all because either the distance comparison (step 2) does not pass or or they are evicted later on (step 3). 
Therefore, instead of collecting vector ids in the top-k structure, we can collect tuples $(k, \mathcal{O})$ recording the current cluster and the offset of the vector in the cluster. 
At the end of the search, when the final top-k results are known, each id can be recovered by looking up the id at offset $\mathcal{O}$ in the id list~$k$.

The wavelet tree (see Section~ \ref{sec:wavelettree}) indexes sequences of symbols. 
For our purpose, we index the sequence of cluster ids $S \in [K)^N$. 
To perform random access of a $(k, \mathcal{O})$ pair, we use a \verb|select| operation on the wavelet tree that returns the index in $S$ of the $\mathcal{O}$\textsuperscript{th} symbol with index $k$: this is the requested id. 
This lookup is performed in logarithmic time.

\subsection{Online setting}\label{sec:online-setting}
In this partial random access setting, we require random access to a specific cluster of vectors at any given time, but not to vectors within a cluster.

\paragraph{IVF indices.}\label{eq:savings-ivf}
For IVF, this implies compression is done per-cluster, with separate bit streams $\bitstream_k \in \{0, 1\}^\star$ for each cluster $k \in [K)$.
Each bit stream stores the subset of ids and codes corresponding to its cluster.
When benchmarking performance of different algorithms we take into account only the total savings resulting from the invariance within clusters, $\sum_{k \in [K)} \log(N_k!)$.

\paragraph{Graph indices.}\label{eq:savings-graphs}
For graphs, there is a separate bit stream $\bitstream_i$ for each node $i \in [N)$, onto which the friend list $\friendlist_i$ is compressed.
Upon visiting the $i$-th node, the search algorithm requires sequential access to codes $\code_j$ for all $j \in \friendlist_i$.
For this reason, we opt to store $\Codes$ uncompressed, to provide random access to such elements.
The compression gains are therefore only from the invariance of friend lists, which total $\sum_{i \in [N)} \log(m_i!)$.

\subsection{Offline setting}
This scenario is equivalent to performing compression of the entire database, where all elements are encoded onto a single bit stream.
The database must be decompressed to perform ANNS, which is motivated by applications where the database needs to be transmitted or stored.

We evaluate only graph-based indices where we compress the entire graph into a single bit stream using Random Edge Coding (REC) \cite{severo2023random} and Zuckerli \citep{versari2020zuckerli}, compression algorithms specifically tailored to compress directed and undirected graphs.

%% file: figs/fig-pq-walltime.tex
\begin{figure*}[ht!]
  \centering
  \begin{minipage}{0.64\textwidth}
    \includegraphics[width=\textwidth]{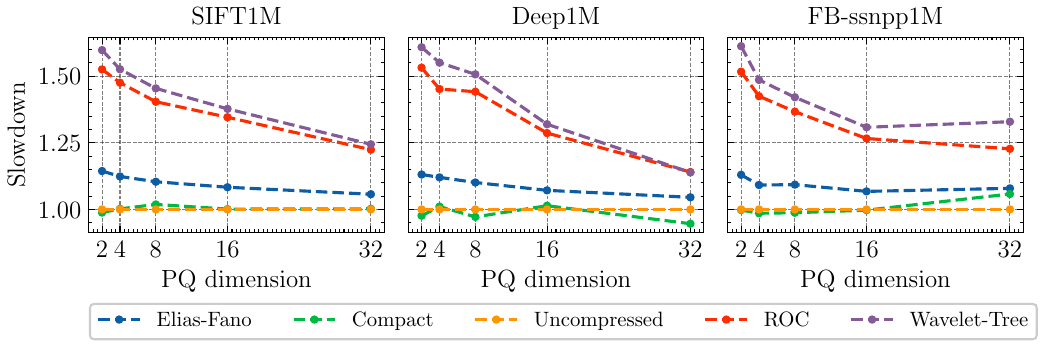}
    \caption{Slowdown relative to the Uncompressed index. As the dimensionality of PQ increases, the overall slowdown is less significant.}
    \label{fig:fig-pq-walltime}
  \end{minipage}\hspace{2em}
  \begin{minipage}{0.24\textwidth}
    \includegraphics[width=\textwidth]{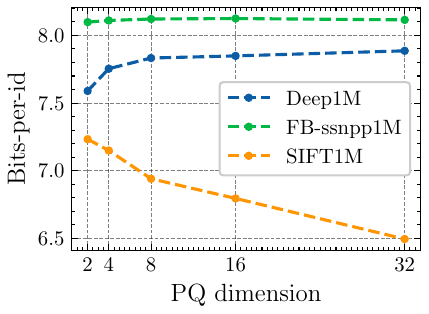}
    \caption{Results for compressing quantization codes conditioned on clusters (originally $8$ bits-per-element). Lower is better.}
    \label{fig:fig-code-compression}
  \end{minipage}
\end{figure*}

%% file: sec/experiments.tex
\section{Experiments}
\label{sec:experiment}

\input{tables/results-online-all}
In this section we discuss our experiments in the different settings of random access. 
We implemented all compression algorithms ($5$ in total) as plugins for the Facebook AI Similarity Search library (Faiss) \cite{douze2024faiss}.
We implemented ROC \cite{severo2022compressing} in C++ for both IVF and NSG indices, and used the REC implementation made available in \cite{severo2023random}. 

\paragraph{Indexing methods.}
FAISS indexes are a combination of a pruning method and a vector compression method. 
We consider the following pruning methods (without compression): IVF$K$ is an IVF structure with $K$ partitions and NSG$L$ is an NSG graph index with at most $L$ outgoing edges per node.
At search time, we fix the number of visited partitions of IVF to 16 and the number of nodes to explore for NSG to 16 as well.
The vector compression is independent of id compression, but it \emph{does} impact the search time (searching compressed vectors is typically slower than uncompressed). 
To account for this, we fix the pruning to IVF1024 and compare several PQ compression settings, where PQ$m$x$b$ means $m$ subquantizers of $b$ bits each ($d=m\times b$, $b$=8 when omitted).

\subsection{Baselines, metrics, datasets}
\label{sec:baselines}
\paragraph{Baselines.}

FAISS stores ids in 64-bits or 32-bit words by default: this is the uncompressed baseline ({\bf Unc.}) 
The compact representation consists in allocating $\lceil \log N \rceil$ bits per id ({\bf Comp.}).
We compare against the classical Elias-Fano (\textbf{EF}) inverted list encoding.
Zuckerli \citep{versari2020zuckerli} is used as a baseline for the offline graph compression setting.

Elias-Fano~\cite{elias1974efficient, fano1971number} encodes the most and least significant bits of the ids in separate bit streams, see \Cref{sec:elias-fano} for details.
We use the optimized Elias-Fano implementation from the Succint library\footnote{\url{https://github.com/ot/succinct/blob/master/elias_fano.hpp}}.
We report the size of the compressed EF state as the sum of both bit streams.

WT1 refers to a WT variant with stronger compression. 
We use the wavelet tree implementation from SDSL lite\footnote{\url{https://github.com/simongog/sdsl-lite}}.

\paragraph{Datasets.}
We experiment on classical datasets for vector search~\cite{pmlr-v176-simhadri22a}, that we restrict to 1 million (1M) vectors for easy comparison. 
The SIFT1M dataset contains 128-dimensional SIFT vectors used for image matching that have a $4\times4\times8$ structure due to their extraction technique~\cite{lowe2004distinctive}. 
The Deep1M and FB-ssnpp are image embeddings~\cite{babenko2016efficient,pizzi2022self}. 
The main difference is that the FB-ssnpp is trained with a loss component whose result is that the transitivity of neighborhoods is hard to use for search.  

\paragraph{Metrics.}
We are interested in speed and memory usage.
Unlike most works on vector search, we do not evaluate the vector search recall, because the compression options we evaluate are lossless so they give the same results.

We report median wall-times, over $100$ runs, for searching with the compressed indices.
All experiments were run on an Intel® Xeon® CPU Processor E5-2698 v4 clocked at 2.20 GHz.
For compression, we report the number of bits per id using the different approaches. 

\subsection{Full and partial random access}
For IVF databases, we applied ROC \cite{severo2022compressing} and Elias-Fano coding \cite{elias1974efficient, fano1971number} to encode the ids of each cluster.
Similarly, we encoded the friend list, $\friendlist_i$, of each node for graph databases.

In each case, the sequence of each cluster or node is encoded into a separate bit stream, to provide partial random access, using the specified compression method.
Results are shown in \Cref{tab:results-online-size}.
For IVFs, we report the bits-per-element (bpe) defined as the sum of bits in all bit streams $\bitstream_k$, $k \in [K)$, divided by the database size ($N=10^6$ for all datasets).
Similarly, for graphs, bpe is the sum of bits in all bit streams $s_i$, $i \in [N)$, divided by the number of edges in the graph.

\paragraph{Main results.}
ROC is able to compress ids to values below the reference, $\ceil{\log(N)} \approx 20$ bits-per-id, for all datasets.
Experiments indicate IVF indices can be compressed down to $14.7\%$ of their original size (i.e., a compression ratio of approximately $7$).
For graph indices the results are less expressive, but still significant, with the best compression rate being to $50\%$ of the original size for NSG256 on FB-ssnpp. %
For NSG16, the result is worse than the $\ceil{\log N}$ baseline, as the friend lists are too short to overcome the initial bits issue %
discussed in \Cref{sec:background-bitsback}.

The wavelet tree (\textbf{WT}) offers full random access to the ids. 
We report two results: 
the baseline WT uses a flat binary binary string, and WT1 uses the RRR structure to compress the strings~\cite{raman2007succinct}. 
WT1 has a higher compression rate at the cost of slower \verb|select| operations.

WT1 is able to surpass the compression rate of ROC in most experiments, at the cost of a significant increase in search time (a factor of $2$ to $3$ times slower when $K \approx \sqrt{N}$).
We conjecture this is due to the data structure exploiting the dependence between lists: together, they form a partition of the interval $[N)$.

\paragraph{Performance improves for larger clusters.}
Compression performance increases with the number of elements in each cluster, or number of indices in each friend list, which can be seen in \Cref{tab:results-online-size}.
For IVF indices, the number of elements in each cluster tends to decrease as the number of clusters increases, since the database size is fixed.

This empirical result is expected, as the number of permutations associated to a set grows with its size.
From \Cref{eq:savings-ivf} we can see that savings are maximized when all elements are in a single cluster, and is equal to $0$ in the extreme case when each element is in its own cluster (i.e., $K = N$).

\paragraph{Optimal compression rates}
For large sets, the difference in bits-per-id between Elias-Fano and the minimal achievable (in the Shannon sense) is known to be constant and roughly $0.56$ bits \cite{elias1974efficient}.
This can be observed in \Cref{tab:results-online-size} by computing the difference between Elias-Fano and ROC for IVF indices, as ROC is close to the Shannon-bound for large sets.

\paragraph{Compute-time}
\Cref{tab:results-online-wall-time} shows the increase in search time as a function of the index size.
For IVF the slowdown in search from using compressed indices is negligible for all compression methods.
The worst case being a $19\%$ slowdown for PQ4 on Deep1M.
Most of the wall-time spent with ROC is due to the Fenwick Tree \cite{fenwick1994} required for entropy coding with ANS \cite{severo2022compressing}.
A future direction is to apply approximate calculations of counts in Fenwick Tree, at the cost of slightly increasing the bits-per-id~\cite{kunzeentropy, kunzepractical}.

The impact of id decoding on runtime is higher relative to the uncompressed case when distance computations are fast. 
The IVF* and NSG* results (9 first rows on the table) are without vector compression, which is efficient, so the impact of id decoding is maximal.
\Cref{fig:fig-pq-walltime} shows the slowdown in search due to id decompression. 
As the dimensionality of PQ increases, so does the distance computation time, which makes the slowdown of id decoding less significant.
Note that PQ8x10 is significantly slower than other variants because the look-up tables used to compute distances are 4x larger. 

\paragraph{Compressing quantization codes.}\label{sec:compressing-codes}
In this section we show that some compression of the codes $\Codes$ is possible when conditioning on cluster indices.
We applied Product Quantization (PQ) \cite{jegou2010product} to an IVF index with $1024$ clusters, and implemented entropy coding with the Craystack \cite{townsend2020hilloc} library for each element of the PQ code, restricted to a single cluster.
More precisely, let $\Codes^{(k)}$ be the matrix of codes for the $k$-th cluster, with 
$\code^{(k, j)} \defeq \left[x_0^{(k, j)}, \dots, x_{N_k-1}^{(k, j)}\right]$
its $j$-th column.
We entropy code each column independently using the distribution that assigns mass to the $i$-th element proportional to the occurrence count up to $i-1$,
\begin{align}
    \Pr&\left(x^{(k, j)}_i = x\;\middle|\;x^{(k, j)}_0, \dots, x^{(k, j)}_{i-1}\right) \\
    & = \frac{1}{256+i}\cdot\left(1 + \sum_{n=0}^{i-1} \mathbf{1}\left[x^{(k,j)}_n = x\right]\right),
\end{align}
for $i > 0$, and equal to $\frac{1}{256}$ for $i=0$ (i.e., the first element).
Initially, this distribution assigns a uniform probability to the occurrence of any byte, and gradually updates the probability as samples are seen.

The entropy of quantization codes $\Codes$ without conditioning on clusters is close to $8.0$ (maximum), implying no further compression is possible. 
Our results for conditional compression are in \Cref{fig:fig-code-compression}.
We are able to compress the codes by up to $19\%$ while still providing random access for SIFT1M, and around $5\%$ for Deep1M.
FB-ssnpp1M does not benefit from further compression.
The compression ratio for SIFT1M is probably due to its structure, that aligns well with PQ sub-vectors. 
The compression improves as the dimensionality of PQ increases.

\subsection{No random access}
\input{tables/results-offline} 
In this setting we investigate compression of graph-based indices.
We applied Random Edge Coding (REC) \cite{severo2023random} and Zuckerli \cite{versari2020zuckerli} to the graph of vertex sets $\Codes$, and friend lists $\friendlist_i$.
Results are shown in \Cref{tab:results-graph-offline}.
For HNSW, we compress only the base level graph since other levels occupy negligible storage. 
The REC algorithm was modified to compress directed graphs by setting $b=0$ in Algorithm 2 in \cite{severo2023random}. 
Compression ratios of up to $2.31$ are achieved with REC.
REC outperforms the Zuckerli baseline in almost all settings, with the exception of HNSW indices for the SIFT1M dataset.
Performance tends to improve for larger graphs, as expected due to savings being proportional to the friend list size (see \Cref{eq:savings-graphs}).
As expected, results improve as the friend lists grow in size.

The number of bits saved from the freedom to re-order the edge list, $\log((\sum_{i \in [N)} m_i)!)$, is significantly larger than that of ROC, $\sum_{i \in [N)} \log\left(m_i !\right)$, resulting in better compression rates.
Second, compressing all edges of the graph into a single ANS state amortizes the initial bits significantly more than compressing each friend list individually.

\subsection{Large-scale experiments}
\input{tables/qinco_results}

We apply id compression to a recent state-of-the-art vector compression method: QINCo~\cite{huijben2024QINco}. 
QINCo, combined with IVF, operates in a high compression regime, so it yields exploitable results with short codes (recall@10=65\% for 8-byte codes).
In that setting, the storage of the ids is important (by default the ids are as large as the vector codes themselves!). 

Table~\ref{tab:qinco} reports search results in this setting, on a billion-scale database with $K=2^{20}$ IVF clusters. 
It shows that the ROC compression reduces the size of the representation by almost 3$\times$, which decreases the overall size of the index from 17.8~GB to 12.5~GB (-30\%).

%% file: tables/results-online-all.tex
\begin{table*}[t]
\centering
\resizebox{\textwidth}{!}{%
\begin{tabular}{lcccggg|cccggg|cccggg}
\toprule
 & \multicolumn{6}{c}{\tablehead{SIFT1M}} & \multicolumn{6}{c}{\tablehead{Deep1M}} & \multicolumn{6}{c}{\tablehead{FB-ssnpp}} \\
& Unc. & Comp. & EF & WT & WT1 & ROC & Unc. & Comp. & EF & WT & WT1 & ROC & Unc. & Comp. & EF & WT & WT1 & ROC \\\midrule
IVF256  & $64$ & $20$ & $9.85$ & $12.1$ & $8.13$ & $9.43$ & $64$ & $20$ & $9.86$ & $12.1$ & $8.42$ & $9.44$ & $64$ & $20$ & $9.87$ & $12.1$ & $8.43$ & $9.46$ \\
IVF512  & $64$ & $20$ & $10.9$ & $13.6$ & $9.23$ & $10.5$ & $64$ & $20$ & $10.9$ & $13.6$ & $9.50$ & $10.5$ & $64$ & $20$ & $10.9$ & $13.6$ & $9.49$ & $10.5$ \\
IVF1024 & $64$ & $20$ & $11.8$ & $15.0$ & $10.3$ & $11.4$ & $64$ & $20$ & $11.9$ & $15.0$ & $10.5$ & $11.5$ & $64$ & $20$ & $11.8$ & $15.0$ & $10.5$ & $11.4$ \\
IVF2048 & $64$ & $20$ & $12.8$ & $16.5$ & $11.3$ & $12.4$ & $64$ & $20$ & $12.9$ & $16.5$ & $11.6$ & $12.5$ & $64$ & $20$ & $12.8$ & $16.5$ & $11.6$ & $12.5$ \\
\midrule
NSG16  & $32$ & $20$ & $18.0$ & - & - & $20.6$ & $32$ & $20$ & $18.0$ & - & - & $20.5$ & $32$ & $20$ & $18.1$ & - & - & $20.9$ \\
NSG32  & $32$ & $20$ & $17.4$ & - & - & $19.4$ & $32$ & $20$ & $17.4$ & - & - & $19.2$ & $32$ & $20$ & $17.5$ & - & - & $19.5$ \\
NSG64  & $32$ & $20$ & $17.3$ & - & - & $18.9$ & $32$ & $20$ & $17.1$ & - & - & $18.6$ & $32$ & $20$ & $16.9$ & - & - & $18.3$ \\
NSG128 & $32$ & $20$ & $17.1$ & - & - & $18.5$ & $32$ & $20$ & $17.0$ & - & - & $18.2$ & $32$ & $20$ & $16.4$ & - & - & $17.2$ \\
NSG256 & $32$ & $20$ & $16.9$ & - & - & $18.0$ & $32$ & $20$ & $16.7$ & - & - & $17.7$ & $32$ & $20$ & $15.7$ & - & - & $16.2$ \\
\midrule
PQ* & $64$ & $20$ & $11.8$ & $15.0$ & $10.3$ & $11.4$ & $64$ & $20$ & $11.9$ & $15.0$ & $10.5$ & $11.5$ & $64$ & $20$ & $11.8$ & $15.0$ & $10.5$ & $11.4$ \\
\bottomrule
\end{tabular}
}
\caption{Compression results for IVF and NSG indices in bits-per-id. All rows, except the last, are for Flat quantizers. Lower is better. Results for the last 3 rows are for an IVF1024 index.
For EF, the sum of bits in both bit streams are reported (without overheads). The Wavelet Tree was not implemented for NSG.
\label{tab:results-online-size}
}
\end{table*}

\begin{table*}[t]
\centering
\resizebox{\textwidth}{!}{%
\begin{tabular}{lcccggg|cccggg|cccggg}
\toprule
 & \multicolumn{6}{c}{\tablehead{SIFT1M}} & \multicolumn{6}{c}{\tablehead{Deep1M}} & \multicolumn{6}{c}{\tablehead{FB-ssnpp}} \\
& Unc. & Comp. & EF & WT & WT1 & ROC & Unc. & Comp. & EF & WT & WT1 & ROC & Unc. & Comp. & EF & WT & WT1 & ROC \\
\midrule
IVF256 & $11.$ & $6.4$ & $11.$ & $6.4$ & $11.$ & $6.4$ & $5.6$ & $5.4$ & $5.4$ & $5.4$ & $4.5$ & $5.4$ & $20.$ & $21.$ & $19.$ & $21.$ & $12.$ & $21.$ \\
IVF512 & $4.1$ & $3.7$ & $4.0$ & $3.8$ & $3.2$ & $3.7$ & $4.5$ & $2.1$ & $2.1$ & $2.3$ & $2.5$ & $2.3$ & $7.0$ & $5.8$ & $5.8$ & $5.9$ & $6.1$ & $5.9$ \\
IVF1024 & $2.6$ & $2.4$ & $2.6$ & $2.5$ & $3.0$ & $2.2$ & $2.8$ & $1.2$ & $1.2$ & $1.3$ & $2.9$ & $1.3$ & $6.8$ & $3.1$ & $4.7$ & $3.2$ & $3.4$ & $3.2$ \\
IVF2048 & $1.9$ & $1.0$ & $2.6$ & $1.1$ & $2.9$ & $1.1$ & $1.6$ & $.69$ & $.72$ & $.75$ & $2.1$ & $.76$ & $2.0$ & $1.8$ & $1.9$ & $1.8$ & $2.4$ & $1.9$ \\
\midrule
NSG16 & $.12$ & $.12$ & $.14$ & - & - & $.29$ & $.19$ & $.20$ & $.22$ & - & - & $.43$ & $.094$ & $.093$ & $.11$ &-  &-  & $.23$ \\
NSG32 & $.26$ & $.21$ & $.24$ & - & - & $.52$ & $.27$ & $.27$ & $.26$ & - & - & $.49$ & $.35$ & $.20$ & $.40$ &  -&  -& $.40$ \\
NSG64 & $.29$ & $.22$ & $.35$ & - & - & $.51$ & $.29$ & $.20$ & $.31$ & - & - & $.53$ & $.54$ & $.74$ & $.57$ &  -&  -& $.89$ \\
NSG128 & $.20$ & $.21$ & $.24$ &-  &-  & $.44$ & $.37$ & $.25$ & $.32$ & - & - & $.54$ & $1.1$ & $1.1$ & $1.1$ &  -&  -& $1.7$ \\
NSG256 & $.27$ & $.28$ & $.32$ &-  &-  & $.54$ & $.22$ & $.24$ & $.27$ & - & - & $.55$ & $2.8$ & $2.8$ & $2.8$ &  -&  -& $4.1$ \\
\midrule
PQ4 & $.33$ & $.31$ & $.33$ & $.36$ & $.75$ & $.37$ & $.31$ & $.28$ & $.30$ & $.35$ & $.74$ & $.35$ & $.36$ & $.25$ & $.36$ & $.29$ & $.81$ & $.28$ \\
PQ16 & $.50$ & $.46$ & $.51$ & $.52$ & $.94$ & $.51$ & $.51$ & $.47$ & $.48$ & $.53$ & $.95$ & $.51$ & $.56$ & $.49$ & $.55$ & $.53$ & $.84$ & $.53$ \\
PQ32 & $.99$ & $.78$ & $1.1$ & $.84$ & $1.6$ & $.84$ & $.85$ & $.92$ & $.92$ & $.98$ & $1.3$ & $.96$ & $.91$ & $.88$ & $.91$ & $.94$ & $1.4$ & $.93$ \\
PQ8x10 & $2.2$ & $2.0$ & $2.2$ & $2.0$ & $2.6$ & $2.0$ & $2.0$ & $1.9$ & $1.9$ & $2.0$ & $2.3$ & $2.0$ & $2.2$ & $2.0$ & $2.1$ & $2.0$ & $2.7$ & $2.0$ \\
\bottomrule
\end{tabular}
}
\caption{Search timings, in seconds, for the compressed and uncompressed indices of \Cref{tab:results-online-size}. Search is performed in parallel on a batch of $10,000$ queries for $\text{nprobe}=16$. Lower is better.
\label{tab:results-online-wall-time}
}
\end{table*}

%% file: tables/results-offline.tex
\begin{table}[]
\centering
\caption{Compression results, in bits-per-id, for Zuckerli \citep{versari2020zuckerli} and Random Edge Coding (REC) \citep{severo2023random}. Lower is better. The Compact reference is $\ceil{\log N} \approx 20$ bits, while Uncompressed is $32$ bits.}
\label{tab:results-graph-offline}
\resizebox{0.95\linewidth}{!}{%
\begin{tabular}{lcg|cg|cg}
\toprule
        & \multicolumn{2}{c}{\tablehead{SIFT1M}}          & \multicolumn{2}{c}{\tablehead{Deep1M}}          & \multicolumn{2}{c}{\tablehead{FB-ssnpp1M}} \\
        & Zuck.            & REC              & Zuck.            & REC              & Zuck.            & REC               \\ \hline
HNSW16  & $17.31$          & $\mathbf{17.29}$ & $17.69$          & $\mathbf{17.23}$ & $17.05$          & $\mathbf{16.08}$  \\
HNSW32  & $\mathbf{15.27}$ & $15.89$          & $\mathbf{15.72}$ & $15.87$          & $14.00$          & $\mathbf{13.60}$  \\
HNSW64  & $\mathbf{14.76}$ & $15.24$          & $15.30$          & $\mathbf{15.24}$ & $14.28$          & $\mathbf{13.88}$  \\
HNSW128 & $\mathbf{14.56}$ & $14.82$          & $15.14$          & $\mathbf{14.81}$ & $14.36$          & $\mathbf{13.87}$  \\
HNSW256 & $\mathbf{14.52}$ & $14.60$          & $15.10$          & $\mathbf{14.58}$ & $14.47$          & $\mathbf{13.84}$  \\ \hline
NSG16   & $\mathbf{17.23}$ & $17.59$          & $\mathbf{17.45}$ & $17.56$          & $16.60$          & $\mathbf{16.18}$  \\
NSG32   & $17.05$          & $\mathbf{16.98}$ & $17.19$          & $\mathbf{16.89}$ & $16.26$          & $\mathbf{15.78}$  \\
NSG64   & $16.93$          & $\mathbf{16.77}$ & $17.04$          & $\mathbf{16.64}$ & $16.00$          & $\mathbf{15.35}$  \\
NSG128  & $16.77$          & $\mathbf{16.60}$ & $16.85$          & $\mathbf{16.47}$ & $15.69$          & $\mathbf{14.88}$  \\
NSG256  & $16.57$          & $\mathbf{16.39}$ & $16.59$          & $\mathbf{16.23}$ & $15.34$          & $\mathbf{14.37}$  \\ \hline
\end{tabular}%
}
\end{table}

%% file: tables/qinco_results.tex
\begin{table}[]
\small
\centering
\caption{
    ID compression rates and times to search 10k vectors in 1B vectors compressed with QINCo. 
    The parameters are set to yield recall@10=0.65: (\texttt{nprobe}, \texttt{quantizer\_efSearch}, \texttt{nshort}) = $(128,64,200)$. 
}
\label{tab:qinco}
\vspace{-0.5em}
\scalebox{0.9}{%
\begin{tabular}{lrrrr}
\toprule
id compression & Unc. & Comp. & EF & ROC \\
\midrule
bits per id & 64      & 30   & 21.81 & 21.46 \\
search time & 11.8    & 11.4 & 11.7  & 14.9  \\
\bottomrule
\end{tabular}%
}
\vspace{-0.5em}
\end{table}

%% file: sec/conclusion.tex
\vspace*{-1.5em}
\section{Discussion and Future Work}\label{sec:conclusion}

This works applies the methods of \citet{severo2022compressing,severo2023random} to ANNS with IVF and graphs indices.
We show a substantial reduction in the bits needed to represent ids is possible, with negligible impact on search speed.
We commit to open-source the code corresponding to this work.

We showed further compression of the quantized codes is possible using ANS when conditioning on the cluster index.
In our experiments, only SIFT1M benefited significantly from further compression, while FB-ssnpp1M and Deep1M did not.
It is expected that codes can be compressed further under distribution shifts of the queries, as redundancies are introduced in the output of lossy compression algorithms (quantizers) in this setting \cite{cover1991, baranchuk2023dedrift}.

REC works by defining a probability model for the sequence of vertices constructed from flattening the edge-list.
Any probability model can be used, as long as it defines a valid conditional distribution over the current vertex given the previous in the sequence.
The model we use~\cite{severo2023random} is tailored to social graphs with power-law degree distributions, which is not the case for NSG nor HNSW graphs.
ROC also makes use of a conditional probability model of the current id given the previous.
For our experiments, we use a uniform model, which does not capture any correlation between ids.

%% file: sec/supplementary.tex
\section{Baseline methods for id encoding}\label{sec:appendix-baselines}

We succintly describe two baseline methods we use to represent unordered sequences of ids: the Elias-Fano coding approach and the Wavelet trees.  

\subsection{Elias-Fano Coding (EF)}\label{sec:elias-fano}

Elias-Fano coding is a compression algorithm for monotone (i.e., non-decreasing) sequences of integers.
For a sequence of length $n$, and maximum value $u$, the integers are first encoded to their binary representations using $\ceil{\log u}$ bits.
Then, EF separately encodes the upper and lower bits of each element into separate bit streams.
The $\ceil{\log u/n}$ lower bits of each element are encoded by concatenating them into a single sequence, in the order they appear in the sequence.
The remaining $\ceil{\log n}$ upper bits are encoded using Unary Coding \cite{cover1999elements}; which can be encoded in at most $2n$ bits.
\footnote{See \url{https://www.antoniomallia.it/sorted-integers-compression-with-elias-fano-encoding.html} for an example and more in detail introduction.}

The optimality of EF is within $0.56$ bits-per-symbol in the following sense.
After encoding, both bit streams together take up roughly $2n + n \log\frac{u}{n}$ (ignoring rounding). Compared to encoding the sequence naively by using $\log u$ bits per symbol, EF provides a saving of $\log \frac{n}{4}$.
The optimal savings per element any algorithm for monotone sequences can offer is $\frac{\log n!}{n}$ \cite{severo2022compressing}, corresponding to the freedom of reordering the sequence normalized by the number of elements. The difference between these quantities converges to approximately $0.56$ for $n \rightarrow \infty$.

\subsection{Zuckerli}
\label{sec:zuckerli}
Zuckerli is a graph compression algorithm that builds off of WebGraph \cite{boldi2004webgraph, boldi2004webgraph2}.
WebGraph partitions each adjacency list such that it matches either 1) a contiguous block of some other adjacency list, 2) a contiguous interval of non-negative integers, or 3) is a sequence of integers that can benefit from delta-encoding.
To perform 1), for each node WebGraph chooses a reference node with a large overlap in the adjacency list.
A sequence of integers for each node is stored, specifying a partition of the reference adjacency list into contiguous blocks.
The protocol specifies that blocks of even index are present in the node's adjacency list, which need not be stored.

Zuckerli improves upon WebGraph by entropy-coding the integer sequence specifying the partition, improving the mechanism that selects the reference node, replacing 2) with run-length encoding \cite{1447423}, as well as other minor improvements.